\documentclass[10pt,twocolumn,letterpaper]{article}

\usepackage{cvpr}              

\definecolor{cvprblue}{rgb}{0.21,0.49,0.74}
\usepackage[pagebackref,breaklinks,colorlinks,allcolors=cvprblue]{hyperref}

\def\paperID{13488} 
\def\confName{CVPR}
\def\confYear{2025}

\title{GIF: Generative Inspiration for Face Recognition at Scale}

\author{Saeed Ebrahimi, Sahar Rahimi, Ali Dabouei,\\ Srinjoy Das, Jeremy M. Dawson, Nasser M. Nasrabadi\\
\tt\small West Virginia University\\ 
\small \{me00018, sr00033, Ad0046\}@mix.wvu.edu, 
\{srinjoy.das, jeremy.dawson, nasser.nasrabadi\}@mail.wvu.edu
}
\usepackage{booktabs} 
\usepackage[skip=2pt]{caption}
\usepackage[ruled,linesnumbered]{algorithm2e}
\usepackage{subcaption}
\usepackage{bm}
\usepackage{graphicx}
\usepackage{enumitem}
\usepackage{multicol}
\usepackage{multirow}
\usepackage{xcolor}
\usepackage{colortbl}
\usepackage{xspace}
\usepackage{amsfonts}
\usepackage{bm}       
\usepackage{diagbox}

\def\bx{\mathbf{x}}
\def\bz{\mathbf{z}}
\def\bw{\mathbf{w}}
\def\bW{\mathbf{W}}
\def\bG{\mathbf{G}}

\def\bH{\mathbf{H}}

\def\bh{\mathbf{h}}

\def\bc{\mathbf{c}}

\def\bz{\mathbf{z}}
\def\bu{\mathbf{u}}

\def\ba{\mathbf{a}}
\def\bb{\mathbf{b}}

\renewcommand{\paragraph}[1]{\par\noindent{\bf #1}}

\DeclareMathOperator*{\minnn}{min}

\usepackage{wrapfig,lipsum,booktabs}
\usepackage{array}
\usepackage{changepage}

\begin{document}
\maketitle
\begin{abstract}
Aiming to reduce the computational cost of Softmax in massive label space of Face Recognition (FR) benchmarks, recent studies estimate the output using a subset of identities.
Although promising, the association between the computation cost and the number of identities in the dataset remains linear only with a reduced ratio. A shared characteristic among available FR methods is the employment of atomic scalar labels during training. Consequently, the input to label matching is through a dot product between the feature vector of the input and the Softmax centroids. Inspired by generative modeling, we present a simple yet effective method that substitutes scalar labels with structured identity code, \ie, a sequence of integers. Specifically, we propose a tokenization scheme that transforms atomic scalar labels into structured identity codes. Then, we train an FR backbone to predict the code for each input instead of its scalar label. As a result, the associated computational cost becomes logarithmic \wrt number of identities.
We demonstrate the benefits of the proposed method by conducting experiments. In particular, our method outperforms its competitors by 1.52\%, and 0.6\% at TAR@FAR$=1e-4$ on IJB-B and IJB-C, respectively, while transforming the association between computational cost and the number of identities from linear to logarithmic. \href{https://github.com/msed-Ebrahimi/GIF}{Code}
\end{abstract}
\vspace{-16pt}    
\section{Introduction}
\label{sec:intro}
\begin{figure}[]
\begin{center}
\includegraphics[width=0.85\linewidth]{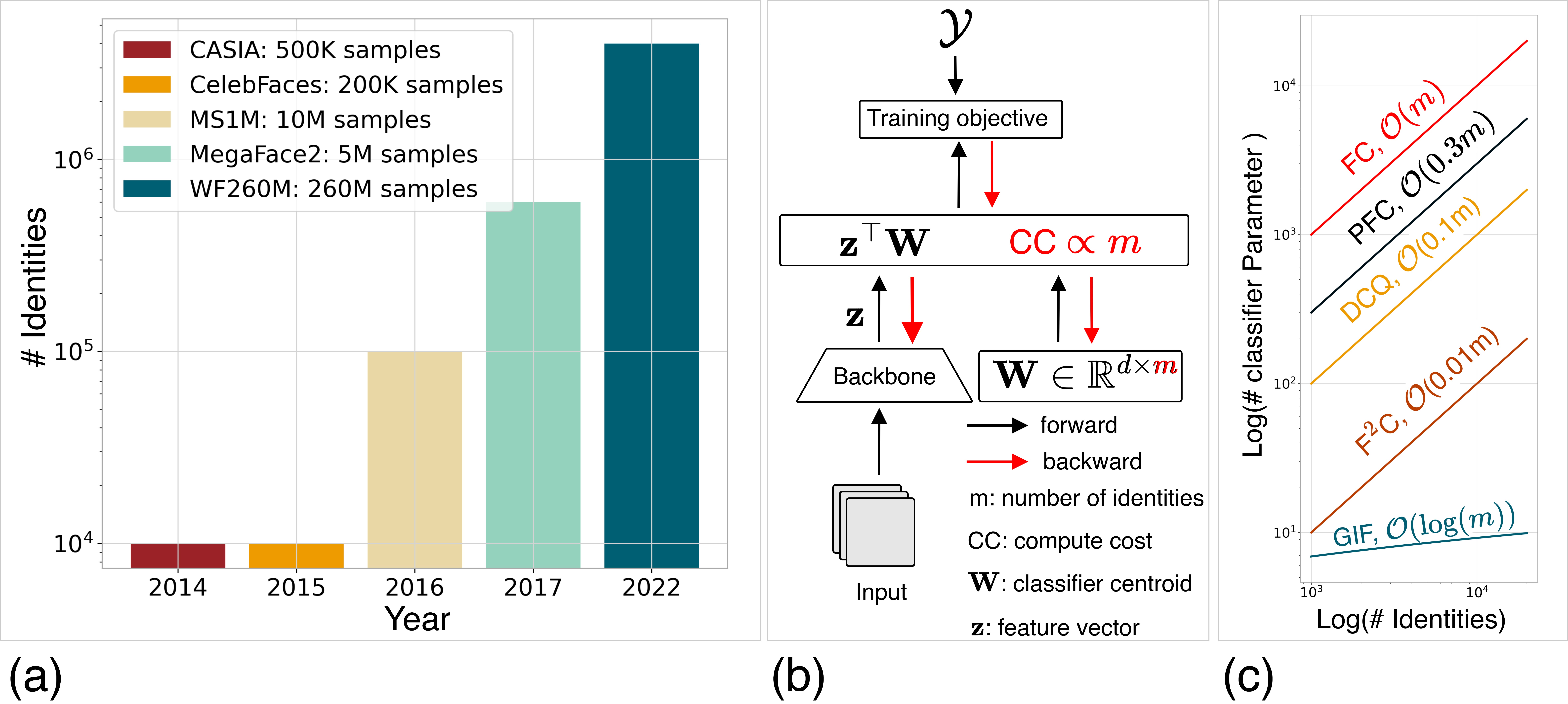}
\end{center}
\vspace{-10pt}
\caption{ a) Illustrating the growth in the number of identities in the FR datasets over time. b) Conventional scalar label leads to the linear association between computational cost and the number of identities.  $\mathcal{D} = \{ (\mathbf{x}_i, y_i) \in \mathcal{X}\times \mathcal{Y}\}$ is the training benchmark, $m$ number of identities, and $\mathcal{V}$ is set of all possible codes. c) Comparing the increase in the computational cost of Fully Connected (FC) \cite{deng2019arcface}, PFC \cite{an2022killing}, DCQ \cite{li2021dynamic}, F$^2$C \cite{wang2022efficient} and GIF as the number of identities increases. GIF significantly reduces the growth rate by changing the scaling from linear to logarithmic.
}\label{fig1}
\vspace{-6mm}
\end{figure}

Angular Margin Softmax (AMS) has been widely used in modern Face Recognition (FR) methods due to its desirable discriminative power and convergence \cite{liu2017sphereface,wang2018cosface,deng2019arcface}. However, AMS training suffers from excessive computational costs in large label spaces \cite{li2021dynamic,an2022killing,shen2023equiangular}. Moreover, the size of FR datasets is steadily increasing in both the number of samples and identities \cite{qiu2021synface,li2021dynamic,wang2022efficient}, shown in Figure \ref{fig1}a. For instance, CASIA-WebFace (2014), consists of 500K samples from 10K identities \cite{yi2014learning}, while WebFace260M (2021), contains 260M samples from 2M identities \cite{zhu2021webface260m}.
\footnote{Commercial datasets are much larger  \cite{schroff2015facenet,li2021dynamic}.}

The common design choice of current FR training associates identities with atomic scalar labels \cite{an2022killing,kim2022adaface,deng2019arcface}. 
Consequently, AMS matches an input to the scalar label by the dot product of the feature vector and AMS centroids, as illustrated in Figure~\ref{fig1}b. This matching framework results in $\mathcal{O}(m)$ computation cost where $m$ is the number of classes \cite{an2022killing,li2021virtual, li2021dynamic, wang2022efficient}.
A potential solution is substituting the AMS with pair-wise cost functions, \eg, contrastive loss, or triplet loss \cite{schroff2015facenet,liu2017sphereface,khosla2020supervised}. However, the combinatorial explosion of the number of possible pairs in the large-scale datasets leads to unstable training and slow convergence \cite{wen2021sphereface2,an2022killing,kim2022adaface,deng2019arcface}. 

The AMS computation cost stems from its normalization over all classes \cite{shen2023equiangular, mettes2019hyperspherical}. Current Efficient Training (ET) methods \cite{an2022killing,li2021virtual, li2021dynamic, wang2022efficient} estimate the AMS output with a portion of classes.
As scalar label setup lacks privileged information, these approaches resort to randomly selecting a subset of identities, leading to suboptimal metric-space exploitation and performance   \cite{mettes2019hyperspherical,luo2018cosine,an2021partial}.
Moreover, the computation cost of the current ET methods still scales linearly with the number of identities, although at a reduced ratio, as shown in Figure \ref{fig1}c. Specifically, the performance of Virtual FC \cite{li2021virtual}, DCQ \cite{li2021dynamic}, F$^2$C \cite{wang2022efficient}, and PFC \cite{an2022killing} peaks at $\mathcal{O}(\alpha m)$, when the $\alpha$, \ie, the parameter defining the portion of classes to select, is 0.01, 0.1, 0.1, and 0.3, respectively.

Current FR methods handle the computational cost of AMS through distributed training on multiple GPUs, \ie, a split of computation on every GPU \cite{an2021partial,ketkar2021introduction,kim2022adaface}. However, the computational constraint remains, and working with limited resources is infeasible. Moreover, large-scale FR benchmarks follow an unbalanced distribution where some identities contain numerous instances, while others have only a few \cite{an2022killing,deng2021variational,zhang2021distribution}. 
In this case, the `\textit{pull-push}' mechanism of the AMS for minor classes is dominated by the pushing force, driving them into a common subspace \cite{deng2021variational,zhu2021webface260m}.
The centroids of the minor classes would be closed or merged, dubbed `\textit{minority collapse}', leading to suboptimal metric space exploitation
\cite{fang2021exploring,deng2021variational,yang2022inducing}. 
Therefore, FR training on large-scale datasets is still far from being solved \cite{li2021virtual}.

\begin{figure}[]
\begin{center}
\includegraphics[width=1.0\linewidth]{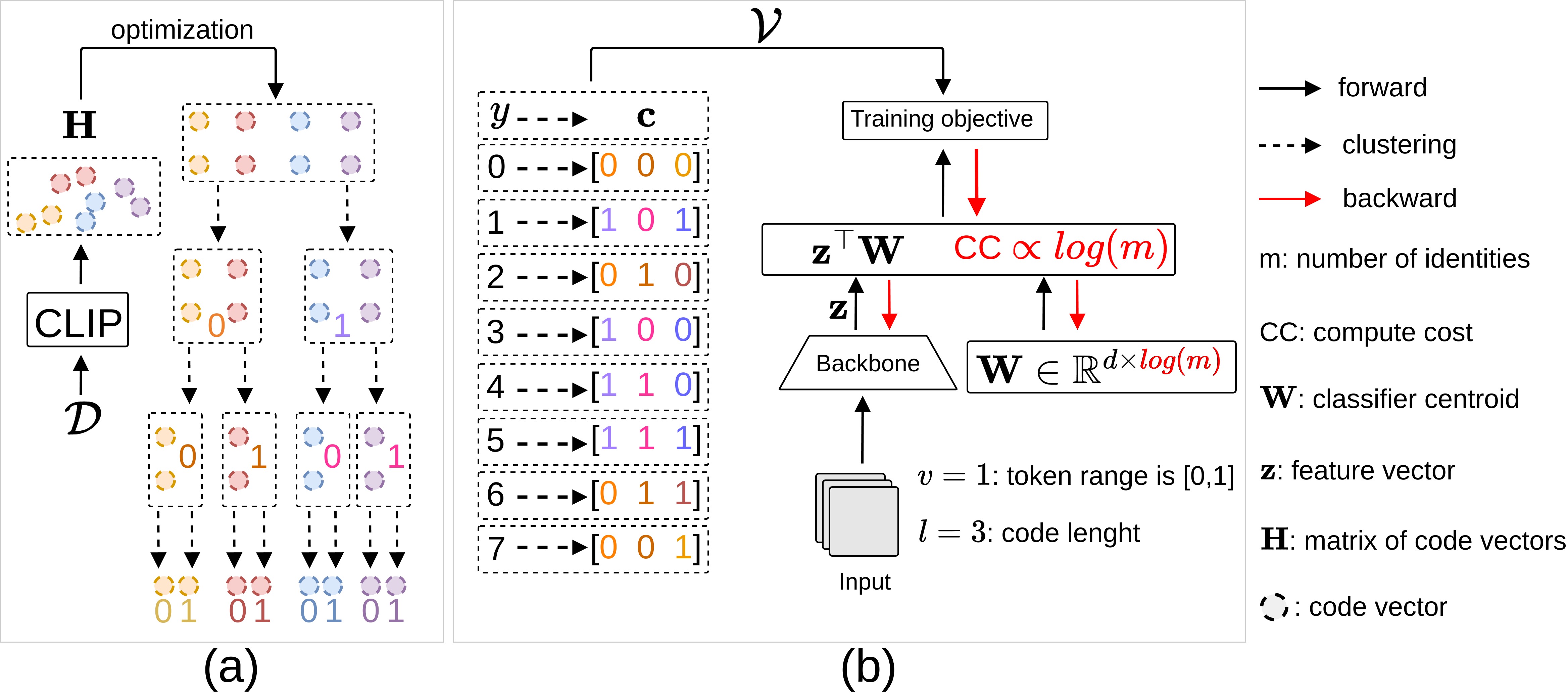}
\end{center}
\caption{a) We first convert scalar labels to identity codes. We use CLIP \cite{radford2021learning} visual encoder to initialize code vectors. Circles with the same colors represent identities with similar generic information. b) Our framework utilizes the identity codes to train the FR model.  $\mathcal{D} = \{ (\mathbf{x}_i, y_i) \in \mathcal{X}\times \mathcal{Y}\}$ is the training benchmark, and $\mathcal{V}$ denotes set of all possible codes. 
}\label{fig2}
\vspace{-6mm}
\end{figure}

Recent advances in large-scale entity recognition and information retrieval have shifted from conventional Softmax classification to generative modeling \cite{de2020autoregressive,caron2024generative,rajput2023recommender}. Specifically, these approaches encode instances as compact sequences of integers, allowing a generative framework to retrieve an instance’s code given a query, rather than directly predicting a class label.
Inspired by this generative modeling, we propose the first large-scale FR training framework that substitutes the scalar labels with identity {codes}, \ie, a sequence of integer tokens.
Consequently, during training, GIF predicts codes' tokens instead of scalar labels.
Note that the number of identities that can be presented using {codes} is the product of the cardinality of each token, \ie, exponential \wrt range of tokens. Therefore, the association between the token range and the number of identities is logarithmic. In this manner,
the GIF formulation changes the linear growth rate of the classifier parameter with the number of classes to logarithmic, \ie, $\mathcal{O}(\log(m))$, as shown in Figure~\ref{fig1}c. 

Constructing the identity codes for FR benchmarks is not trivial since 1) no privileged information is accessible, and 2) unstructured (atomic) codes are inadequate in large-scale classification \cite{de2020autoregressive,kudo2018sentencepiece,hu2023open, tay2022transformer,caron2024generative}.
We circumvent this issue with \textit{a priori} approximation of the unit hypersphere with a semantically structured and discriminative set of code vectors. Then, we construct the identity {codes} by applying the hierarchical clustering on the code vectors, as shown in Figure~\ref{fig2}a.
Our proposal decomposes FR training into two independent steps: 1) identity tokenization and 2) FR training. This decoupling alleviates the `\textit{minority collapse}' issue of conventional FR since the organization of code vectors is independent of a number of per-class instances. The main contributions of this paper can be summarized as follows:

\begin{itemize}
    \item We propose an FR training framework that facilitates training on large-scale benchmarks. Specifically, we formalize FR training so that the linear growth rate of the computational cost with the number of classes, \ie, $\mathcal{O}(m)$, changes to logarithmic, \ie, $\mathcal{O}(\log(m))$. 
    
    \item We propose a strategy for converting FR scalar labels into structured identity codes in which identities with similar generic information share some code tokens.

    \item We show that the proposed method resolves the `\textit{minority collapse}' problem of the conventional classification framework given an unbalanced dataset.
    
\end{itemize}

\section{Related Works}
\label{sec:RW}
\subsection{Face Recognition at Scale}
FR benchmarks have been growing in the number of identities and samples \cite{wang2018devil,guo2016ms,nech2017level,bansal2017umdfaces,parkhi2015deep,cao2018vggface2,yi2014learning,wang2017normface,zhu2021webface260m}. Specifically, early FR benchmarks consisted of thousands of identities/samples, while current benchmarks contain millions of identities/samples, as shown in Figure \ref{fig1}a. Large-scale FR benchmarks and Angular Margin Softmax (AMS) empower the FR community to train high-performance models \cite{zhu2021webface260m,kim2022adaface,deng2019arcface}. However, integrating the AMS classifier with million-scale identities, \ie, classes, results in huge computational complexities \cite{an2022killing,wang2022efficient,li2021dynamic,li2021virtual}. 
Specifically, computing the AMS output involving matrix multiplication between the feature vector and AMS centroids \cite{kim2022adaface, liu2017sphereface, wang2018cosface}.
Consequently, as the number of identities increases, this process becomes prohibitively expensive \cite{an2022killing,an2021partial,wang2022efficient}. 

An intuitive solution is to substitute the AMS with pair-wise supervision \cite{schroff2015facenet,liu2017sphereface,khosla2020supervised}. However, the combinatorial explosion of possible pairs in large-scale datasets leads to unstable training and convergence issues \cite{sun2018multi,wen2021sphereface2,an2022killing,saadabadi2023quality,kim2022adaface,deng2019arcface}. Recent studies try to reduce the computational cost through approximating AMS output with a subset of classes \cite{an2021partial,zhang2018accelerated,li2021virtual,li2021dynamic}. Zhang \etal \cite{zhang2018accelerated} propose partitioning the AMS centroids using a hashing forest and employing the classes within for every sample to compute the output.
An \etal \cite{an2021partial} randomly samples a portion of classes to compute the output. Li \etal \cite{li2021virtual} randomly split label spaces into groups. Each group shares the same anchor, which is used in AMS computation. Li \etal \cite{li2021dynamic} abandoned learnable centroids and utilized a momentum-encoder \cite{he2020momentum} to compute centroids. Although promising, the growth rate of computational cost with the number of identities remains linear. Moreover, employing a limited number of negative classes leads to suboptimal performance \cite{wang2022efficient,li2021dynamic, mettes2019hyperspherical,an2022killing}.

\section{Method}
\subsection{Notation}
Let $\mathcal{D} = \{ (\mathbf{x}_i, y_i) \in \mathcal{X}\times \mathcal{Y}\}_{i=1}^{n}$ be the training dataset, consisting of $n$ faces from $m$ identities.
In this work, we substitute a scalar label $y_i$ by the {code} $\mathbf{c}^{y_i} = \{c^{y_i}_1,\dots, c^{y_i}_l\} \in [ 0, v-1 ]^{(l)}$, \ie, a sequence of integers. The variable $l$ denotes the length of the {code} and $v$ is the range of all integer values that each {code} token $c^{y_i}_j$ takes.
This forms the codes' vocabulary $\mathcal{V}$ with up to $|\mathcal{V}|=v^{(l)}$ unique {codes}, \ie, the total number of identities that $\mathcal{V}$ can present.
Note that the vanilla scalar label fits into our {code} formulation when $l=1$ and $v = m$, \ie, the {codes} will be equivalent to the scalar labels.
$\bH =  [\bh_0,\dots,\bh_{m-1}] \in \mathbb{R}^{d\times m} $ is the matrix storing code vectors. 
$F_{\theta}$ denotes a deep neural network with trainable parameter $\theta$ that maps an input face $\mathbf{x} \in \mathbb{R}^{3\times h \times w}$, to a $d$-dimensional representation $\mathbf{z} = F_{\theta}(\bx) \in \mathbb{R}^d$. ${\mathbf{W}}=[\bw_{{1}},\bw_{2}, \dots,\bw_{m}] \in  \mathbb{R}^{d\times m}$ is the matrix storing Softmax centroids $\bw_i \in \mathbb{R}^{d}$.
For convenience of presentation, all representations are $\ell_2$-normalized.

\subsection{Problem Definition}\label{PD}
Current state-of-the-art (SOTA) FR training frameworks employ the AMS for end-to-end training \cite{liu2017sphereface,wang2018cosface,deng2019arcface}:
\begin{equation}\label{bilevel1}
\resizebox{0.8\linewidth}{!}{$
\begin{aligned}
  \mathop{{\minnn}}_{{\theta,\bW}} \sum_{i=1}^{n}L_{CE}(\overline{y}_i,y_i); \quad   \overline{y}_i = \frac{e^{s \langle \mathbf{w}_{y_i}, F_{\theta}({\bx_i}) \rangle}}{\sum_{k=1}^{m} e^{s \langle \mathbf{w}_{k}, F_{\theta}({\bx_i}) \rangle}}
,
\end{aligned}$}
\end{equation}
where $L_{CE}$ represents the Cross-Entropy (CE) loss, $s$ is introduced as the scaling hyper-parameter which affects the curves of the output \cite{zhang2019adacos}, and $\langle \mathbf{a},\mathbf{b} \rangle$ denotes the cosine of the angle between two vectors $\mathbf{a}$ and $\mathbf{b}$. Here, we remove angular margins for notational convenience. 
The global normalization of Equation~\ref{bilevel1}, \ie, multiplication between $\bz_i$ and $\bw_k$ in the denominator, leads to $\mathcal{O}(m)$ computation cost.

Existing methods \cite{an2022killing,wang2022efficient,li2021dynamic,li2021virtual} reduce this computational load by approximating $\overline{y}$ with a subset of identities. However, the association between computational cost and $m$ remains linear, only with a reduced ratio, \ie, $\mathcal{O}(\alpha m)$ where $0<\alpha<1$. 
Moreover, real-world FR benchmarks follow an unbalanced distribution where some identities have plenty of instances, while others have only a few, \cite{saadabadi2023quality,deng2021variational,zhu2021webface260m}. 
In this scenario, AMS optimization, \ie, Equation~\ref{bilevel1}, suffers from `\textit{minority collapse}'. Please see Section \ref{cederivative} in Supplementary Material for a detailed analysis. 

\subsection{Overview.} GIF substitutes scalar labels with structured identities {codes}. In this way, instead of solving an $m$-way classification problem, GIF solves $l$ parallel $v$-way classifications. Note that $v^l \geq m$ reflects the total number of classes that can be presented using {codes}, \ie, logarithmic association between $v$ and number of classes $v \propto \log(m)$. Therefore, $m$-way classification of GIF leads to a logarithmic scaling of computational cost with the number of identities. 

It is not trivial to convert scalar labels to identity codes as 1) the atomic code fails in large-scale classification \cite{tay2022transformer,mettes2019hyperspherical,caron2024generative}, and 2) no privileged information is accessible in FR \cite{mettes2019hyperspherical,rawat2017deep}. 
We circumvent these issues with a two-stage approach: 1) assigning each identity to a code vector on the unit-hypersphere $\mathcal{S}^{d-1}$, \ie, $y_i \rightarrow \bh_{y_i}$, and 2) assigning each $\bh_{y_i}$ to a {code} in $\mathcal{V}$, \ie, $\bh_{y_i} \rightarrow \mathbf{c}^{y_i}$. 
We note that the consistency between dataset semantics, \ie, inter-class relations, and the organization of code vectors is crucial for structured code.
Therefor, we follow \cite{mettes2019hyperspherical,mikolov2013distributed}, and employ the CLIP \cite{radford2021learning} visual encoder to initialize $\bH$, \ie, $y_i \rightarrow \bh_{y_i}$.

Although CLIP embedding provides semantic consistency between $\bH$ and $\mathcal{D}$, it lacks the inter-identity separation.
We follow recent works on hyperspherical metric-space \cite{hoffer2018fix,mensink2013distance,movshovitz2017no,mettes2019hyperspherical,wang2022visual,shen2023equiangular} to enhance inter-identity separation among code vectors: in hyperspherical embedding, the uniform distribution of classes enhances inter-class separation and metric-space exploitation. 
We optimize $\bH$ to follow a uniform distribution over $\mathcal{S}^{d-1}$ in the sense that each pair of the $\bh_{y_i}$ and $\bh_{y_j}$ is maximally separated. 
We formulate this optimization independent of the $\mathcal{D}$, circumventing the `\textit{minority collapse}' issue.

To construct the identity {codes}, we form a tree structure by applying hierarchical clustering over code vectors $\bH$, as shown in Figure~\ref{fig2}a. The {code} for each identity $\bc^{y_i}$ is the concatenation of the node indices $c_j^{y_i}$ along the path from the root to a corresponding leaf node, \ie, $\bh_{y_i}\rightarrow \bc^{y_i}$. In this manner, faces with similar generic information have some overlapping {code} tokens $c_j$, \ie structured identity {codes}. 
Please note that we obtain identity codes before the main training for a given $\mathcal{D}$. Then, we use codes to train an arbitrary FR backbone on the $\mathcal{D}$. Code vectors $\bh_{y_i}$ and identity codes $\bc^{y_i}$ are fixed during the training
and GIF FR training solely involves an optimization problem \textit{w.r.t.} model parameters. Hence, our main objective is:
\begin{equation}\label{bilevel2}
 \small
 \begin{aligned}
  \mathop{{\minnn}}_{{\theta,\phi_1,\dots,\phi_l}} \sum_{i=1}^{n}L(F_{\theta}(.), H_{\phi_{1\leq j\leq l}}(.),\bx_i, \bc^{y_i}),
\end{aligned}
\end{equation}
where $L$ represents the learning objective function, and $H_{\phi_j}$ is the classifier for $l$-th token with trainable parameter $\phi_j$.

\begin{figure*}[t]
\centering
\includegraphics[width=1.0\linewidth]{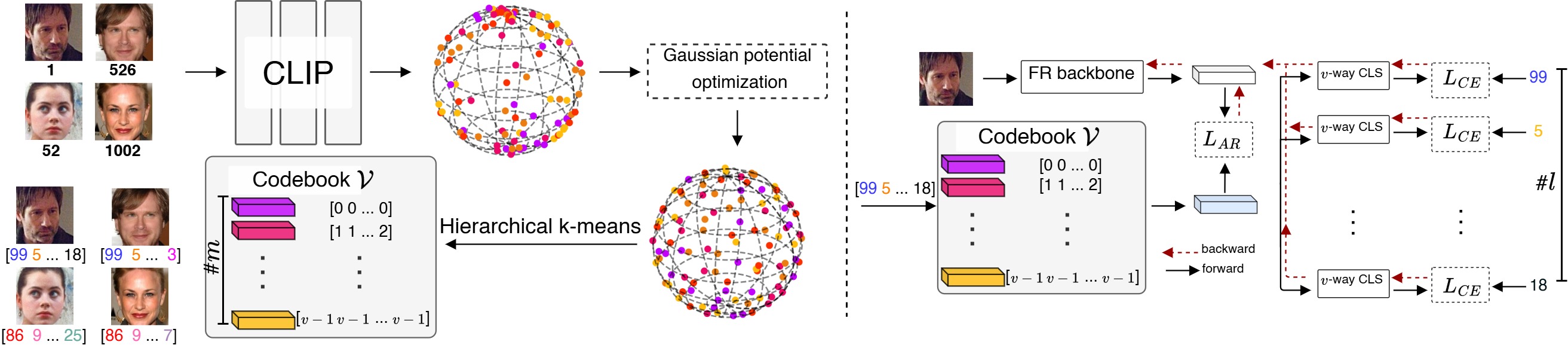}
\caption{Left) Overview of proposed tokenization. We position each $y_i$ on the $\mathcal{S}^{d-1}$ using hyperspherical code vectors $\bh_{y_i}$ in a way that the pair-wise distance among arbitrary $\bh_{y_i}$ and $\bh_{y_j}$ is maximized. Then, we construct identity codes by applying hierarchical clustering over $\bH$. Right) Overview of proposed FR training pipeline.}\label{mainfig}
\vspace{-16pt}
\end{figure*}

During training, GIF predicts the identity {code} of the $\bx_i$ by solving $l$ parallel $v$-way classification.
In this way, GIF circumvents the requirement for computing global normalization over $m$ entities. Instead, our method computes normalization over $v \propto \log(m)$ possible token values. Furthermore, the distribution of code vectors $\bh_{y_i}$ is fixed before training independently of the number of samples associated with each identity. This effectively circumvents the `\textit{minority collapse}' problem in the unbalanced distribution sample per-classes in FR benchmarks.
Figure \ref{mainfig} and Algorithm \ref{alg1} provide an overview of the proposed method.

\subsection{Substituting Scalar Labels with Identity Codes}
The intuitive way of creating {codes} is to tokenize the text assigned to entities by $\mathcal{D}$, \ie, either classes' name, caption, or description \cite{de2020autoregressive,hu2023open,kudo2018sentencepiece,mettes2019hyperspherical}.
In this way, each token value maps to a sub/word in a pre-defined vocabulary, then the entity codes correspond to the tokenized text used in the dataset to describe class $y_i$  \cite{kudo2018subword, sennrich2015neural,wang2022git,kudo2018sentencepiece}.
However, in FR, such privileged information is inaccessible and inadequate, rendering the employment of text tokenizers impractical. 
Here, we detail the proposed tokenization scheme that operates independently of privileged information.

\subsubsection{Structured and Separable Code Vectors}
In FR, tokenization using scalar labels is infeasible since scalar labels lack inter-identity relations. 
We circumvent this issue by a tow steps tokenization scheme: 1) assigning each scalar label to a hyperspherical code vector ($y_i \rightarrow \bh_{y_i}$), and 2) constructing identity codes from code vectors ($\bh_{y_i} \rightarrow \bc^{y_i}$).
It is essential that the organization of code vectors reflects dataset semantics, \ie, inter-identity relations. 
To inject dataset semantics into the distribution of $\bh_i$, we leverage the CLIP \cite{radford2021learning} visual embedding as the initialization for $\mathbf{H}$. Formally, each $\mathbf{h}_{y_i}$ is initialized by averaging over the CLIP representations from 
class $y_i$:
\begin{equation}\label{clipinit}
\resizebox{0.45\linewidth}{!}{$
\begin{aligned}
\mathbf{h}_{y_i} = \frac{1}{|\mathcal{D}_{y_i}|}\sum_{\mathbf{x}\in\mathcal{D}_{y_i}} CLIP(\mathbf{x}),
\end{aligned}$}
\end{equation}
where $\mathcal{D}_{y_i}$ refers to all the samples from identity $y_i$ . 
$\bH$ obtained from Equation~\ref{clipinit} provides semantic alignment; however, it lacks adequate inter-identity separation.

Studies \cite{kasarla2022maximum,duan2019uniformface,mettes2019hyperspherical,wang2020understanding,yang2022inducing,yaras2022neural} have shown that uniformly distributed hyperspherical points enhance metric-space exploitation and inter-entity separation.
For ${d=2}$ and ${m}$ points, this problem reduces to splitting the circle into equal slices with \(\dfrac{2\pi}{m}\) angles. 
However, no optimal solution exists in \(d \!\! \geq \!\! 3\) \cite{saff1997distributing, kong2018recurrent}\footnote{This is known as the Tammes problem \cite{tammes1930origin}.}. 
Inspired by the recent progress in hyperspherical representation learning \cite{wang2020understanding,khosla2020supervised,caron2020unsupervised}, we employ a metric based on the Gaussian potential kernel to encourage uniform distribution of code vectors over ${{S}^{d-1}}$. 
Let ${p(.)}$ be the distribution over ${{S}^{d-1}}$. {${G_t(\mathbf{a},\mathbf{b})\triangleq S^{d-1} \!  \times \!  S^{d-1}\rightarrow \mathbb{R}_{+}}$} is the Gaussian potential kernel: 
\begin{equation}\label{GPK}
\begin{aligned}
G_t(\ba,\bb)\triangleq e^{-t||\ba-\bb||_2^2}; \quad t>0,
\end{aligned}
\end{equation}
the uniformity loss can be defined as:
\begin{equation}\label{ALGPK}
\begin{aligned}
	L \triangleq \log{\mathop{\mathbb{E}}_{\bh_i,\bh_j \mathop{\backsim}\limits^{\text{i.i.d.}} p_{\bh}}}[G_t(\ba,\bb)];\quad t>0,
\end{aligned}
\end{equation}
 which is nicely tied with the uniform distribution of points on the $\small{S^{d-1}}$ \cite{bochner1933monotone}; please refer to \cite{wang2020understanding} for detailed derivative.

We use iterative optimization to minimize Equation \ref{ALGPK}. Specifically, we optimize a subset of $\mathbf{H}$ at every iteration:
\begin{equation}\label{practical_loss}
 \small
 \begin{aligned}
	L_{GP} = \log{(\frac{1}{\hat{m}} \sum_{i=1}^{\hat{m}}\sum_{j=1}^m {g_{i,j}})},
\end{aligned}
\end{equation}
where $\small{\hat{m} < m}$, \ie, a subset of $\bH$ is randomly selected, and  $\small{g_{i,j}}$ is the element of $\small{\bG}$ that reflects the pairwise Gaussian potential between $\small{\bh_i}$ and $\small{\bh_j}$. 
Equation~\ref{practical_loss} merely depends on the metric space dimension, $d$, and the number of classes, $m$. 
Therefore, the distribution of the resulting $\bH$ is not affected by the unbalanced distribution of samples in $\mathcal{D}$, and avoids `\textit{minority collapse}'.

\subsubsection{Constructing Identity {Codes} from Code Vectors}
Each entity in $\bH$, \ie, code vector, refers to an identity of $\mathcal{D}$. To construct codes from $\bH$, we apply the hierarchical $k$-means algorithm over $\bH$. Given the instances of $\bH$ to be indexed, all $\bh_i$ are first assigned into $k$ clusters using their angular similarity, \ie, cosine. Here, the $k$ in clustering refers to the token range $v$.
For each subsequent token $c_j$, where $2 \leq j \leq l-1$, the $k$-means algorithm is applied recursively on every cluster. This iterative division results in a hierarchical token structure where each cluster at token position $j$ encapsulates no more than $v^{l-j}$ entities. Finally, for the $l$-th token, each element is assigned an arbitrary number from $[0,v-1]$.
In this way, we organize all identities into a tree structure.
Each identity is associated with one leaf node with a deterministic routing path $\bc^{y_i}=\{c_1^{y_i},\dots, c_{l}^{y_i}\}$ from the root. Each $c_j \in [0, v-1]$ represents the internal cluster index for level $j$, and $c_{l} \in [0, v-1]$ is the leaf node. The code for an identity is the concatenation of the node indices along the path from the root to its corresponding leaf node. 

\begin{figure}[t]
\centering
\includegraphics[width=1.0\linewidth]{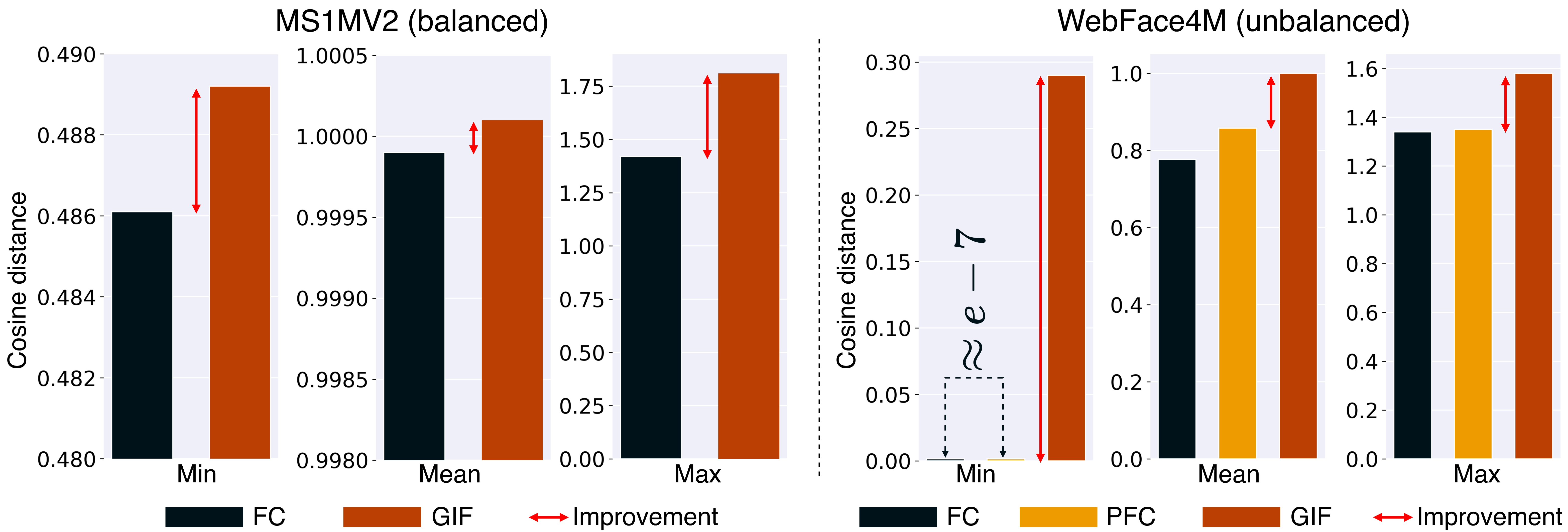}
\caption{The maximum ${\uparrow}$, minimum ${\uparrow}$, and mean ${\uparrow}$ pairwise cosine distance among Softmax centroids of Fully Connected (FC) ArcFace (FC) \cite{deng2019arcface}, PFC \cite{an2022killing}, and the code vectors $\bh_i$ of our proposal when $d=512$. More separation among $\bh_i$ reflects better metric-space exploitation, leading to more discrimination power in the embedding of $F_{\theta}$. The separation of Softmax centroids is based on the final training checkpoint.}\label{InterClassSeparation}
\vspace{-16pt}
\end{figure}

\subsection{The GIF Training}
Our proposal substitutes scalar labels with identity {codes}.
Consequently, the training objective of GIF is to predict tokens of identity {codes} given the input face image:
\begin{equation}\label{bilevel4}
\resizebox{0.9\linewidth}{!}{$
\begin{aligned}
  &L_C(\bx_i,\bc^{y_i}) =\sum_{j=1}^{l}\lambda_j  L_{CE}(\overline{c}^{y_i}_j, c^{y_i}_j),\\
  &\overline{\bc}^{y_i}=[\overline{c}_1^{y_i},\dots, \overline{c}_l^{y_i} ]; \quad \overline{c}_j^{y_i} = \frac{e^{\gamma \langle \mathbf{u}_{c^{y_i}_j},H_{\phi_j}(F_{\theta}({\bx_i})) \rangle}}{\sum_{k=0}^{v-1} e^{\gamma \langle \mathbf{u}_{c_j^k}, H_{\phi_j}(F_{\theta}({\bx_i})) \rangle}},
\end{aligned}$}
\end{equation}
where $\lambda_j$ balances the contributions of the each token, $H_{\phi_j}(.)$ is the projection head with trainable parameter $\phi_j$ corresponding to $j$-th token and ${\mathbf{U}^j}=[\bu^j_{{0}},\dots,\bu^j_{v-1}] \in  \mathbb{R}^{d\times v}$ is its Softmax classifier. 
Comparing Equations~\ref{bilevel4} and \ref{bilevel1}, 
the single normalization factor with $m$ operation is changed to $l\ll m$ normalization factors with $v \propto \log m$ operations. This significantly reduces training memory costs and simplifies distributed implementation.

\begin{algorithm}[t]
\scriptsize
\caption{GIF}\label{alg1}
Initialize $F_{\theta}$, $H_{\phi_i}\:\: \forall i\in[1,l]$, $\gamma > 0 $,  $\lambda_i >0 \:\:\forall i\in[1,l]$, $t_1>0$ and $t_2>0$

{\tcc{{\textcolor{blue}{Start Tokenization}}}}

\For{${y=0}$ \KwTo ${m-1 }$}{  
Initialize $\bh_y \in \bH$ using Equation \ref{clipinit}  \CommentSty{ \textcolor{blue}{$\triangleright$ $y_i \rightarrow \bh_{y_i}$}}
}
\For{$\small{t=0}$ \KwTo $\small{t_1\!-\!1 }$}{  

Compute $\small{L_{uni}}$, $\small{\forall \bh_i \in {\bH}}$ 

Update $\small{{\bH}}$ using $\small{\nabla_{{\bH}}L_{uni}}$ \CommentSty{ \textcolor{blue}{$\triangleright$ optimizing code vectors}}
}
\For{$j = 1$ \KwTo $l-1$}{

    \ForEach{cluster from the $(j-1)$-th level }{
        Divide into $v$ sub-clusters \CommentSty{\textcolor{blue}{$\triangleright$ Recursive clustering}}
        
    }
}

\For{ $\bh_{y_i}$ in ${\bH}$}{
    $\bc^{y_i} \gets []$ 
    
    \For{$j = 1$ \KwTo $l$}{ 
        $c_j^{y_i} =$ cluster index of $\bh_{y_i}$ at level $j$
        
        Append $c_j^{y_i}$ to $\bc^{y_i}$ 
    } 
    \CommentSty{\textcolor{blue}{$\triangleright$ $\bh_{y_i} \rightarrow \bc^{y_i}$}}
}
\tcc{{\textcolor{blue}{End Tokenization }}}
\tcc{{\textcolor{blue}{Main training}}}

\For{${t=0}$ ... $\small{t_2\!-\!1 }$}{
\For{Batch in $\mathcal{D}$}{
 ${\bz={F_{\theta}}(Batch)}$

 $\overline{\bc}^{y_i} = [H_{\phi_1}(\bz),\dots, H_{\phi_l}(\bz)]$
 
Compute $L$ using Equation \ref{mainobj}

Update ${{F_{\theta}}}, H_{\phi_1},\dots,H_{\phi_l}$} 
}

\end{algorithm}

Although efficient, solely optimizing Equation~\ref{bilevel4} does not explicitly encourage the intra-class compactness property in the $F_{\theta}$ embedding.
Thus, we employ regression supervision, which directly encourages the intra-class compactness of $F_{\theta}$. Specifically, for a training sample $\bx_i$, we seek to learn a mapping from the input to its assigned code vector, \ie, `\textit{pull}' $\bz_i=F_{\theta}(\bx_i)$ toward $\bh_{y_i}$:
\begin{equation}\label{cosineSimilarity}
 \small
 \begin{aligned}
 L_{AR}(\bx_i, \bh_{y_i}) = \frac{1}{2} {(\bz_i^\top\bh_{y_i}-1)^2},
\end{aligned}
\end{equation}
this loss directly encourages the alignment between $\small{\bz_i}$ and $\small{\bh_{y_i}}$, supervising the intra-class compactness by decreasing the angular distance of samples and their assigned $\bh_{y_i}$. 

Equation \ref{cosineSimilarity} is only concerned with aligning the samples to their assigned prototype.
Specifically, the partial derivative of Equation \ref{cosineSimilarity} can be given as:
\begin{equation}\label{gardCosSim}
 \small
 \begin{aligned}
 \frac{\partial L_{AR}}{\partial \bz_i} = - (1-\bz_i^\top\bh_{y_i})\bz_i^\top\bh_{y_i},
\end{aligned}
\end{equation}
which is identical to the `\textit{pull}' force in the CE derivative. Please see Section \textcolor{cvprblue}{6} in Supplementary Material for detailed derivative of CE. 
Thus, Equation \ref{cosineSimilarity} avoids the `\textit{minority collapse}' issue since there is no `\textit{push}' force in its derivative. 
Moreover, the same global minimizer, \ie, equiangular organization of  $\bH$ and $\bW$, holds for both Equations \ref{bilevel1} and \ref{cosineSimilarity}. Please refer to \cite{yang2022inducing} for detailed derivative.
Finally, the main GIF training objective is:
\begin{equation}\label{mainobj}
\begin{aligned}
  \mathop{{\minnn}}_{{\theta,\phi_1,\dots,\phi_l}} \sum_{i=1}^{n}\left[L_C(\bx_i,\bc_i) + \gamma L_{AR}(\bx_i,\bh_{y_i})\right],
\end{aligned}
\end{equation}
where $\gamma$ balances the contribution of each loss to the training. This optimization implicitly and explicitly encourages the discriminative power of the $F_{\theta}$ embedding.

\section{Experiment}

\subsection{Datasets}
We utilize the cleaned versions of WebFace260M \cite{zhu2021webface260m}, \ie, WebFace42M (42M images from 2M identities), WebFace12M (12M images from 600K identities) and WebFace4M (4M images from 200K identities), as training sets. For evaluations, we employ the standard academic benchmarks of LFW \cite{huang2008labeled}, CPLFW \cite{zheng2018cross}, CALFW \cite{zheng2017cross}, CFP-FP \cite{sengupta2016frontal}, AgeDB \cite{moschoglou2017agedb}, IJB-B \cite{whitelam2017iarpa}, and IJB-C \cite{maze2018iarpa}.
As per the conventional FR framework, all datasets used in our work are aligned and transformed to $\small{112 \times 112}$ pixels.

\subsection{Implementation Details}
To initialize $\bH$, we employ CLIP image features to obtain per-class mean representation.
For the optimization of $\bH$, \ie, minimizing Equation \ref{practical_loss}, we used SGD optimizer with a constant learning rate of 0.1 for 1000 epochs with the {batch size of 2K on each GPU.
The hyperparameter of $\gamma$ governing the balance between $L_C$ and $L_{AR}$ is set to one, and $\lambda_i$ is set to one for all $i\in[1,l]$. Furthermore, $l$ in different training datasets is set in a way that $5\leq v\leq 20$. Section \ref{abblambda} in Supplementary Material provides detailed experiments on these parameters.
We construct identity codes by applying the hierarchical k-means algorithm on the optimized $\bH$.
For the backbone, we employ ResNet-100 \cite{deng2019arcface,he2016deep} and ViT-base architectures \cite{dosovitskiy2020image}. Moreover, $H_{\phi}$ is a small MLP consisting of two hidden layers with size $d$ and ReLU activation.
When the backbone is ResNet, we use SGD optimizer, with a cosine annealing learning rate starting from 0.1 for 20 epochs with a momentum of 0.9 and a weight decay of 0.0001.
We employ the AdamW optimizer, which has a base learning rate of 0.0001 and a weight decay of 0.1, to train ViT for 40 epochs using a 512 batch size on each GPU.
All experiments utilize eight {Nvidia A100}.

\begin{table}[]
\addtolength{\tabcolsep}{-3pt}
\resizebox{1.0\linewidth}{!}{
\begin{tabular}{l|c|c|c|ccccc}
\toprule
Method     & Train Set  & $F_{\theta}$ & $\mathcal{O}(.)$  & LFW   & CFP-FP & AgeDB  &  IJB-B & IJB-C  \\ \midrule
Virtual FC \cite{li2021virtual}& MS1MV2     & R100    & $\frac{1}{100}m$ & 99.38 & 95.55  &   -    &  -      & -       \\
DCQ \cite{li2021dynamic}       & MS1MV2     & R100    & $\frac{1}{10}m$ & 99.80 & 98.44  & 98.23 &  -          & -           \\
F$^{2}$C  \cite{wang2022efficient}      & MS1MV2     & R50     &$\frac{1}{10}m$ & 99.50 & 98.46  & 97.83  &      -      & 94.91 \\
   \rowcolor[HTML]{C0C0C0} GIF        &    MS1MV2        &   R100      & $\log m$ &    \textbf{99.85}   &   \textbf{98.80}     &      \textbf{98.58}        & \textbf{95.05}           &  \textbf{96.77}           \\ \midrule
PFC \cite{an2022killing}   & WebFace4M  & R100    & $\frac{3}{10}m$ & \textbf{99.85} & 99.23  & 98.01       & 95.64 & 97.22  \\
     \rowcolor[HTML]{C0C0C0} GIF     &     WebFace4M       &    R100     & $\log m$&   \textbf{99.85}    &    \textbf{99.36}    &       \textbf{98.55}      &    \textbf{96.90}        &         \textbf{97.83}    \\ \midrule
PFC  \cite{an2022killing}  & WebFace12M & R100    & $\frac{3}{10}m$& 99.83 & 99.40  & 98.53       & 96.31 & 97.58  \\
  \rowcolor[HTML]{C0C0C0} GIF        &       WebFace12M      &      R100   & $\log m$& \textbf{99.85}      &  \textbf{99.46}      &      \textbf{98.81}        &      \textbf{97.08}      &            \textbf{97.82} \\ \midrule
F$^{2}$C \cite{wang2022efficient}       & WebFace42M & R100    & $\frac{1}{10}m$ & 99.83 & 99.33  & 98.33  &    -        &       -      \\
PFC \cite{an2022killing}   & WebFace42M & R100    & $\frac{3}{10}m$& 99.85 & 99.40  & 98.60       & 96.47 & 97.82 \\ 
  \rowcolor[HTML]{C0C0C0}  GIF       &    WebFace42M        &   R100      & $\log m$&   \textbf{99.85}    &   \textbf{99.80}     &      \textbf{99.40}      &     \textbf{97.99}       &  \textbf{98.42}  \\ \midrule
PFC \cite{an2022killing}  & WebFace42M & ViT    & $\frac{3}{10}m$& \textbf{99.83}  & 99.40  &  98.53      & 96.56 & 97.90 \\ 
  \rowcolor[HTML]{C0C0C0} GIF         &       WebFace42M & ViT        & $\log m$&   \textbf{99.83}    &     \textbf{99.48}   &        \textbf{96.16}    &       \textbf{97.24}     &  \textbf{97.99}  \\ 
           
           \bottomrule        
\end{tabular}}
\caption{{Performance comparison with SOTA ET approaches. Verification accuracy (\%) is reported for LFW, CFP-FP, and AgeDB. TAR@FAR$=1e-4$ is reported for IJB-B and IJB-C.}} 
\label{EfficientFR}
\vspace{-16pt}
\end{table}

\subsection{Hyperspherical Separation}\label{sep}

Separation among $\bh_i$ affects the discrimination power of our final FR model since 1) $\bh_i$ remain fixed during the training, and 2) $\bh_i$ are explicitly and implicitly employed to train $F_{\theta}$.
Figure~\ref{InterClassSeparation} compares inter-class separation of FC \cite{deng2019arcface}, and PFC centroids \cite{an2022killing}, with the proposed code vectors, \ie, min, max, and mean pair-wise cosine distance \cite{mettes2019hyperspherical}.  
The entities in $\bH$ show superior separation scores across datasets with different numbers of identities. These results showcase the generalizability of using the Gaussian potential to encourage separability across ranges of $m$. It is worth noting that since the embedding dimension is conventionally constant across different FR methods \cite{schroff2015facenet,deng2019arcface,kim2022adaface,an2022killing}, \ie, $d=512$, Figure~\ref{InterClassSeparation} does not illustrate the results when $d$ changes. 

Moreover, results in Figure~\ref{InterClassSeparation} empirically illustrate the `\textit{minority collapse}' issue when the FR dataset follows an unbalanced distribution. Specifically, a near zero minimum pair-wise distance among the centroids of the FC and PFC on WebFace4M shows that at least two classes in their set of centroids are merged or extensively close to each other.
Our proposal resolves this issue by eliminating the effect of per-identity samples in $\bh_i$ optimization.
Specifically, our proposed optimization for $\bH$
operates independently of the distribution of samples across classes. Instead, it merely depends on the metric-space dimension, $d$, and the number of classes, $m$. Please note MS1MV2 follows a balanced distribution of samples across classes with an average of 100 images for each identity \cite{li2021dynamic,deng2019arcface}.

\subsection{Comparison with SOTA Approaches}
To demonstrate the effectiveness of GIF, we conduct evaluations with two sets of approaches: 1) Efficient Training (ET), and 2) Conventional Distributed Training (CDT).

\noindent\textbf{Comparison with ET:}
Table~\ref{EfficientFR} illustrates that GIF outperforms prior ET methods across different datasets with identities ranging from 85K to 2M. Notably, GIF improves ET baselines on LFW, CFP-FP, and AgeDB even though the performance on these celebrity benchmarks tends to be saturated.
Using MS1MV2 and R100, GIF improves prior methods by remarkable margins of 0.34\%, and 0.35\% on CFP-FP and AgeDB, respectively. Using variants of WebFace, GIF enhances all previous ET methods across LFW, CFP-FP and AgeDB evaluations.
The GIF enhancements extend beyond large-scale or unbalanced training benchmarks, including both balanced (\eg, MS1MV2) and unbalanced (\eg, WebFace), as well as large-scale datasets (\eg, WebFace12M and WebFace42M). 
Importantly, GIF obtains these improvements while significantly reducing the computational burden of prior approaches, underscoring its capability to generalize and effectively exploit metric-space in real-world FR applications.

Moreover, using R100 as the backbone and WebFace4M, WebFace12M, and WebFace42M as the training data, GIF surpasses its competitors by considerable margins of 1.26\%, 0.77\%, and 1.52\% at FAR=$1e-4$ on IJB-B, respectively.
Consistently, employing MS1MV2, WebFace4M, WebFace12M, and WebFace42M training sets and R100 as the backbone, GIF outperforms prior ET approaches by large margins of 1.86\%, 0.61\%, 0.24\%, and 0.6\% at TAR@FAR=$1e-4$ on IJB-C, respectively. These consistent advancements with different training datasets on challenging IJB evaluations showcase that GIF can be effectively scaled to real-world FR training frameworks. Moreover, GIF outperforms PFC, \ie, previous SOTA ET, across ResNet-100 and ViT-base networks, using WebFace42M as a training dataset. Notably, GIF outperforms PFC with a considerable margin of 0.68\% and 0.1\% in IJB-B and IJB-C evaluations at TAR@FAR$=1e-4$, respectively. These improvements across different network architectures showcase the generalization of GIF across different backbones.

\begin{table}
\addtolength{\tabcolsep}{-5pt}
\resizebox{1.0\linewidth}{!}{
\begin{tabular}{l|c|ccccccc}
\toprule
Method               & Train Set & LFW   & CPLFW & CALFW & CFP-FP & Age-DB & IJB-B & IJB-C \\ \midrule
CosFace \cite{wang2018cosface}            & MS1MV2                & 99.81 & 92.28 & 95.76 & 98.12  & 98.11  & 94.80 & 96.37 \\
ArcFace \cite{deng2019arcface}            & MS1MV2             & 99.83 & 92.08 & 95.45 & 98.27  & 98.28  & 94.25 & 96.03 \\
GroupFace \cite{kim2020groupface}         & MS1MV2             & 99.85 & 93.17 & 96.20 & 98.63  & 98.28  & 94.93 & 96.26 \\
DUL \cite{chang2020data}                & MS1MV2              & 99.83 & -     & -     & 98.78  & -      & -     & 94.61 \\
CurricularFace \cite{huang2020curricularface}      & MS1MV2            & 99.80 & 93.13 & 96.20 & 98.37  & 98.32  & 94.80  & 96.10 \\
BroadFace \cite{kim2020broadface}          & MS1MV2         & 99.85 & 93.17 & 96.20 & 98.63  & 98.38  & 94.97 & 96.38 \\
MagFace \cite{meng2021magface}            & MS1MV2            & 99.83 & 92.87 & 96.15 & 98.46  & 98.17  & 94.51 & 95.97 \\
  \rowcolor[HTML]{C0C0C0} GIF   &   MS1MV2            &    \textbf{99.85}   & \textbf{94.45}      &  \textbf{96.94}     &  \textbf{98.80}      &     \textbf{98.58}   &   \textbf{95.05}    &   \textbf{96.77}    \\ \midrule
ArcFace \cite{deng2019arcface}            & WebFace4M     & 99.83 & 94.35 & 96.00 & 99.06  & 97.93  & 95.75 & 96.63 \\
         \rowcolor[HTML]{C0C0C0}GIF     & WebFace4M             &  \textbf{99.85}     &    \textbf{95.03}   & \textbf{96.85}      & \textbf{98.36}       &  \textbf{98.55}      &   \textbf{96.90}    & \textbf{97.83}      \\ \midrule
  AdaFace  (ViT) \cite{kim2022adaface}          & WebFace4M     & 99.80 & 94.97 & 96.03  & 98.94   &97.48   & 95.60 & 97.14 \\
         \rowcolor[HTML]{C0C0C0}GIF (ViT)                & WebFace4M     &    \textbf{99.83}   &  \textbf{95.97}     &  \textbf{97.19}     & \textbf{99.69}       &    \textbf{98.19}    &  \textbf{96.68}     &  \textbf{97.92}     \\       \bottomrule
\end{tabular}}
\caption{Performance comparison to SOTA FR training approaches. Verification accuracy (\%) is reported for LFW, CFP-FP, and AgeDB. TAR@FAR$=1e-4$ is reported for IJB-B and IJB-C.} \label{conventionalFR}
\vspace{-10pt}
\end{table}

\noindent{\textbf{Comparison with CDT:}}
Table~\ref{conventionalFR} compares GIF with SOTA CDT methods across varying dataset sizes. 
Using MS1MV2 as the training set, GIF improves previous CDT approaches across all evaluations. Moreover, employing WebFace4M dataset and ResNet-100, GIF outperforms ArcFace with 1.15\%, and 1.2\% at TAR@FAR$=1e-4$ on IJB-B, and IJB-C, respectively. Concretely, with WebFace4M as the training set and ViT-base backbone, GIF supersedes AdaFace with a considerable margin of 1.08\%, and 0.78\% at TAR@FAR$=1e-4$ on IJB-B, and IJB-C, respectively.
These enhancements across diverse evaluation metrics, training benchmarks, and backbones underscore the efficacy of GIF in metric-space exploitation.

The improvements on WebFace4M are more significant than on MS1MV2. Both datasets are semi-constrained \cite{liu2022controllable}, but MS1MV2 maintains a balanced distribution of samples per class. The unbalanced sample distribution in WebFace4M more clearly highlights the benefits of our method over prior approaches. Specifically, previous works suffer from `\textit{minority collapse}', while GIF circumvents this issue by decoupling the optimization of code vectors from the number of per-class instances. Moreover, GIF improves over its competitors with significantly less computational cost, underscoring its efficacy in developing discriminative power using structured identity code vectors.

\begin{figure*}[]
\begin{center}
\includegraphics[width=1.0\linewidth]{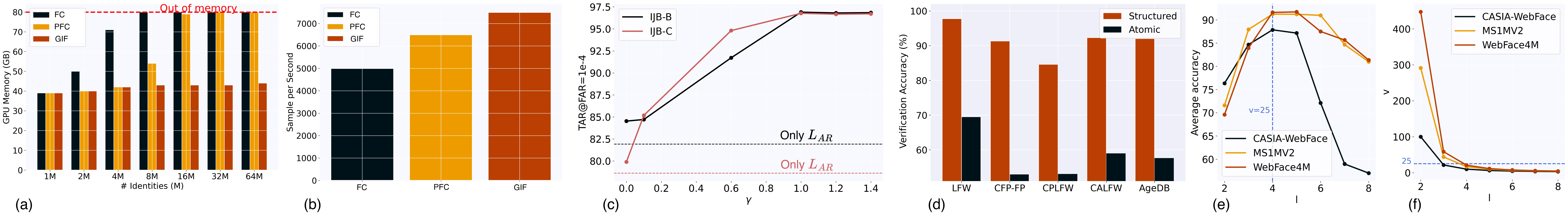}
\end{center}
\vspace{-5pt}
\caption{
(a, b) GPU memory consumption and Training speed comparison between FC, PFC and GIF: GIF significantly improves training cost compared to both FC \cite{deng2019arcface} and PFC \cite{an2022killing}. (c) Ablation on loss components when the training data is MS1MV2, backbone is ResNet-100. (d) Comparing the evaluation performance between employing structured vs. atomic identity codes during training: training using atomic code fails. (e) Effect of the length of codes in the performance when the dataset and label space changes. The average performance across LFW, CFP-FP, CPLFW, CALFW, and AgeDB is reported. (f) The range of tokens, \ie, $v$, when the $l$ changes across datasets.
}\label{Abb1}
\vspace{-3mm}
\end{figure*}

\subsubsection{Training Cost and Efficiency}
To empirically study the GPU memory consumption of GIF, we conduct experiments using ResNet-100 as the backbone. Each GPU processes a batch size of 512, and the number of identities changes from 1M to 64M. As illustrated in Figure~\ref{Abb1}a, the benefits of GIF become more evident as the number of identities increases.
Specifically, when the number of identities reaches 8M \cite{schroff2015facenet,li2021dynamic}, conventional FC layers lead to an out-of-memory (OOM) error, even with distributed training across eight Nvidia A100 80G GPUs. 
PFC \cite{an2022killing} consumes less memory by utilizing a fraction of classes, \ie, \(\frac{3}{10}m\).
However, when the number of identities exceeds 16M, the PFC cannot be implemented on the aforementioned machine.
GIF framework significantly decreases the training memory cost with almost negligible additional memory consumption when the number of identities ranges from 1M to 64M.

Moreover, Figure~\ref{Abb1}b compares the training speed of GIF with FC \cite{deng2019arcface} and PFC \cite{an2022killing}, using the WebFace42M as training dataset and ResNet-100 as the backbone. Results show that GIF significantly enhances training speed, achieving more than 20\% and 15\% improvements over FC \cite{deng2019arcface} and PFC \cite{an2022killing}, respectively. These substantial improvements in training speed and reductions in memory consumption, along with enhanced performance as demonstrated in Tables \ref{EfficientFR} and \ref{conventionalFR}, highlight our method's efficacy for large-scale face recognition. 
For a detailed analysis of the computational costs of optimizing code vectors, see Section \ref{ccc} in the Supplementary Material.

\subsection{Ablations}
\subsubsection{Loss Components}
Here, we investigate the effect of each component of Equation~\ref{mainobj}, \ie, the main FR training objective, on the performance of GIF. To this end, we conduct a series of ablation studies on IJB-B and IJB-C datasets using MS1MV2 training set and ResNet-100, as shown in Figure~\ref{Abb1}c.
The results illustrate the benefits of adding explicit embedding supervision, \ie, $L_{AR}$, to the code prediction loss, \ie, $L_{C}$. 
These improvements showcase the efficacy of our approach to organizing semantically structured and well-separated code vectors on the unit hypersphere.
Figure \ref{Abb1}c shows that balancing the coefficient of $L_C$ and $L_{AR}$ results in the best performance, \ie, $\gamma = 1$. We attribute this to 1) the fine-grain nature of FR, which requires explicit and implicit supervision over embedding in the GIF framework, and 2) the complementary role of $L_{AR}$ to $L_{C}$.
Additionally, Figure~\ref{Abb1}c also compares scenarios where training relies solely on $L_C$ ($\gamma = 0$) or $L_{AR}$ (indicated by horizontal dashed lines). In both IJB-B and IJB-C evaluations, $L_C$ consistently outperforms $L_{AR}$. These results confirm that open-set FR is too complex for the backbone to merely learn embeddings via angular regression loss.

\subsubsection{Structured vs. Atomic Codes}
Here, we compare the performance of GIF when using randomly constructed identity {codes}, \ie, atomic codes, with the structured {codes}. 
To purely investigate the effect of {codes}, we solely used $L_{C}$ as the training signal in the experiments of this section.
To construct atomic {codes}, we randomly pick one code for every scalar label from the set of all possible codes. 
Results in Figure~\ref{Abb1}d show that even in datasets with almost saturated performance, \ie, LFW, CFP-FP, CPLFW, CALFW, and Age-DB, atomic {codes} struggles to perform slightly better than the random guess. Given FR benchmarks' massive label space and FR open-set nature, these results align with the findings of \cite{mettes2019hyperspherical, caron2024generative}. We note that initializing code vectors from a random distribution instead of using the CLIP \cite{radford2021learning} visual encoder is the same as having atomic codes, \ie, applying k-means on the random discrete approximation of hypersphere results in atomic codes.

\subsubsection{Length of {Codes} and Range of Tokens}
Here, we explore the effect of $l$, \ie, the length of identity codes, when the number of identities changes. To this end, we run experiments using three different datasets of CASIA-WebFace, MS1MV2, and WebFace4M consisting of 10k, 85k, and 200k identities, respectively. 
To solely investigate the effect of the {codes} length, we only used $L_C$ as the training signal. 
Figure \ref{Abb1}e shows that employing two tokens results in suboptimal performances across datasets. We attribute this to the inadequacy of tokenization in capturing the dataset's hierarchical structure using $l=2$, making the model's learning more challenging.

Figure \ref{Abb1}e shows that the performance of GIF increases drastically from $l=2$ to $l=4$ across all training benchmarks. Concretely, Figure \ref{Abb1}f illustrates that with $l=4$, the range of the tokens is $v<=25$.
These experiments empirically demonstrate that tokenization best captures the hierarchical structure of current public training benchmarks when the length of codes is chosen so that each token is from $[0,25]$.
Furthermore, as expected, increasing the $l$ after some points results in significant performance degradation, \ie, $l=6$. Large $l$ results in small clusters densely scattered in the embedding. Therefore, the learning for the model becomes more challenging and causes convergence issues.  
Based on these observations, in different datasets with varying sizes of identities, \ie, $m$, we choose $l$ in a way that  $v_l<v<v_h$ where $v_l=5$, and $v_h=25$. 

\subsubsection{Effect of Separability of Code Vectors in FR}

\begin{table}[]
\addtolength{\tabcolsep}{3pt}
\resizebox{1.0\linewidth}{!}{
\begin{tabular}{l|cccccc}
\toprule
\multirow{2}{*}{Method} & \multicolumn{3}{c}{IJB-B} & \multicolumn{3}{c}{IJB-C} \\
                        & 1e-4    & 1e-3   & 1e-2   & 1e-4    & 1e-3   & 1e-2   \\ \midrule
GIF-CLIP                & 24.30   & 44.83  & 69.66  &  29.62  &  50.63 & 73.32  \\
  GIF                     & 95.05   & 97.04  & 98.02  & 96.77   & 97.17  & 97.79  \\ 
  \rowcolor[HTML]{C0C0C0} Improvement & \textbf{70.75}   & \textbf{52.21} & \textbf{28.36}  & \textbf{67.15}   & \textbf{46.54} &\textbf{24.47} \\ \bottomrule
\end{tabular}}
\caption{Abblation on the distribution of $\bH$. Performance using ResNet-100 as the backbone and MS1MV2 as the training data. Naively employing $\bH$ initialized from CLIP lacks the discriminative power required for FR.} \label{AbbCLIP}
\end{table}

Table~\ref{AbbCLIP} compares FR performance when GIF uses optimized code vectors and when it uses the initial per-identity mean CLIP embeddings.
These results underscore the importance of integrating the notion of separability into the code vectors organizations. Specifically, the optimized \( \mathbf{H} \) markedly enhances performance across datasets. 
The results from initialized code vectors are much better than the random guess, \eg, nearly 0.01\% TAR@FAR$=1e-4$, in IJB-B. This shows that, although CLIP embeddings provide a reasonable level of identity similarity, they lack the discriminatory power necessary for FR.

\section{Conclusion}
In this paper, we proposed an FR training framework based on predicting identity codes, \ie, a sequence of integers, to address the prohibitive computational complexities of current SOTA FR approaches. Our method converts the scalar labels of current FR benchmarks into identity codes, \ie, tokenization, without requiring privileged information. The proposed tokenization scheme constructs a structured identity code in which faces with similar general information share some code tokens. Significantly, our formulation changes the linear association of FR training computation requirement with a number of classes $\mathcal{O}(m)$ into a logarithmic relation $\mathcal{O}(\log(m))$. It is worth noting that the proposed method reduces the computational cost of the current FR method without sacrificing the FR performance and outperforms its competitors across different evaluations. The efficacy of the proposed method is evaluated through experiments across diverse training benchmarks.
{
    \small
    \bibliographystyle{ieeenat_fullname}
    \bibliography{main}
}
\clearpage
\setcounter{page}{1}
\maketitlesupplementary

\section{CE Derivative}
\label{cederivative}

Considering layer-peeled model to make a tractable analysis \cite{yang2022inducing,fang2021exploring}, the gradient of Equation \ref{bilevel1} \wrt the $\bw_j$ is:
\begin{equation}\label{classifierGrad}
\resizebox{0.9\linewidth}{!}{$
\begin{aligned}
 \frac{\partial L_{CE}}{\partial \bw_j}\!\! =\!\! \sum_{i=1}^{n}\left[{-(\!1\!-\!p_j(\bz_i)){\bz_i} \delta(j,y_i)}+ {{p_j(\bz_i)}{\bz_i}(\!1\!-\!\delta(j,y_i))}\right],
\end{aligned}$}
\end{equation}
here $p_j(\bz)$ is the predicted probability that $\bz=F_{\theta}(\bx)$ belongs to the $j$-th class and $\delta(i,j)$ is one if $i$ is equal to $j$ and 0 otherwise. 
We can reformulate the Equation~\ref{classifierGrad} to the following form:
\begin{equation}\label{classifierGrad2}
\begin{aligned}
 -\frac{\partial L_{CE}}{\partial \bw_j} = \mathbf{f}_{\rm{pull}}+\mathbf{f}_{\rm{push}},
\end{aligned}
\end{equation}
where $\mathbf{f}_{\rm{pull}}^{(\bw_j)}=\sum_{i=1}^{n^+}[\left(1-p_j\left(\bz_i\right)\right)\bz_i]$, $\mathbf{f}_{\rm{push}}^{(\bw_j)}  =-\sum_{i=1}^{n^-}p_j[\left(\bz_{i}\right)\bz_{i}]$, $n^+$ represents samples belonging to the $j$-th class, \ie, positives, and $n^-$ denotes the samples from other classes, \ie, negatives.
Equation~\ref{classifierGrad2} reveals that CE pulls $\bw_j$ toward the positive instances, \ie, $n^+$, while pushing $\bw_j$ away from negative ones, \ie, $n^-$.

Large-scale FR benchmarks follow an imbalanced distribution where some identities have plenty of instances, while others only contain a few, \ie, $n^+ \ll n^-$ \cite{saadabadi2023quality,deng2021variational,zhu2021webface260m}.
Consequently, the optimization of $\bw_j$ of minority classes is predominantly influenced by $\mathbf{f}_{\rm{push}}$.
Additionally, $\mathbf{f}_{\rm{push}}$ is approximately uniform across all minority classes and forces them to the same subspace \cite{deng2021variational,yang2022inducing}. Thus, centroids of minority classes merge, \ie, dubbed `\textit{minority collapse}', lowering inter-class discrimination and metric-space exploitation. Despite the progress of available methods \cite{an2022killing,li2021dynamic}, the `\textit{minority collapse}' issue remains unsolved as long as identity centroids' optimization remains dependent on the number of per-identity instances.

\section{Ablation on $\lambda$}
\label{abblambda}
Here, we examine the impact of varying each \(\lambda_j\) on the training process. Each \(\lambda_j\) quantifies the relative importance of the \(j\)-th token during training, where a higher \(\lambda_j\) indicates greater impact of the corresponding token, and a lower \(\lambda_j\) suggests less importance.
We conducted a series of ablation studies using the IJB-B and IJB-C datasets with the WebFace4M training set and a ResNet-100 backbone. Our objective was to isolate the influence of the length of the codes; hence, we utilized only \(L_C\) as the training signal. We explored four distinct patterns in the distribution of \(\lambda_i\): increasing, decreasing, Gaussian, and uniform, as depicted in Figure \ref{abblambda}a.
For each configuration, we normalized the \(\lambda_i\) values so that their sum equals one. The results, shown in Figure \ref{abblambda}b, indicate that a uniform of \(\lambda\) across token indices yields superior performance compared to non-uniform. This finding aligns with our identity tokenization strategy, wherein the search space is sequentially narrowed with each correctly predicted token \(c^{y_i}_j\) of the identity code \(\mathbf{c}^{y_i}\). Based on these findings, we employed a uniform \(\lambda = \frac{1}{l}\) in our experiments.

\begin{figure}[t]
\centering
\includegraphics[width=1.0\linewidth]{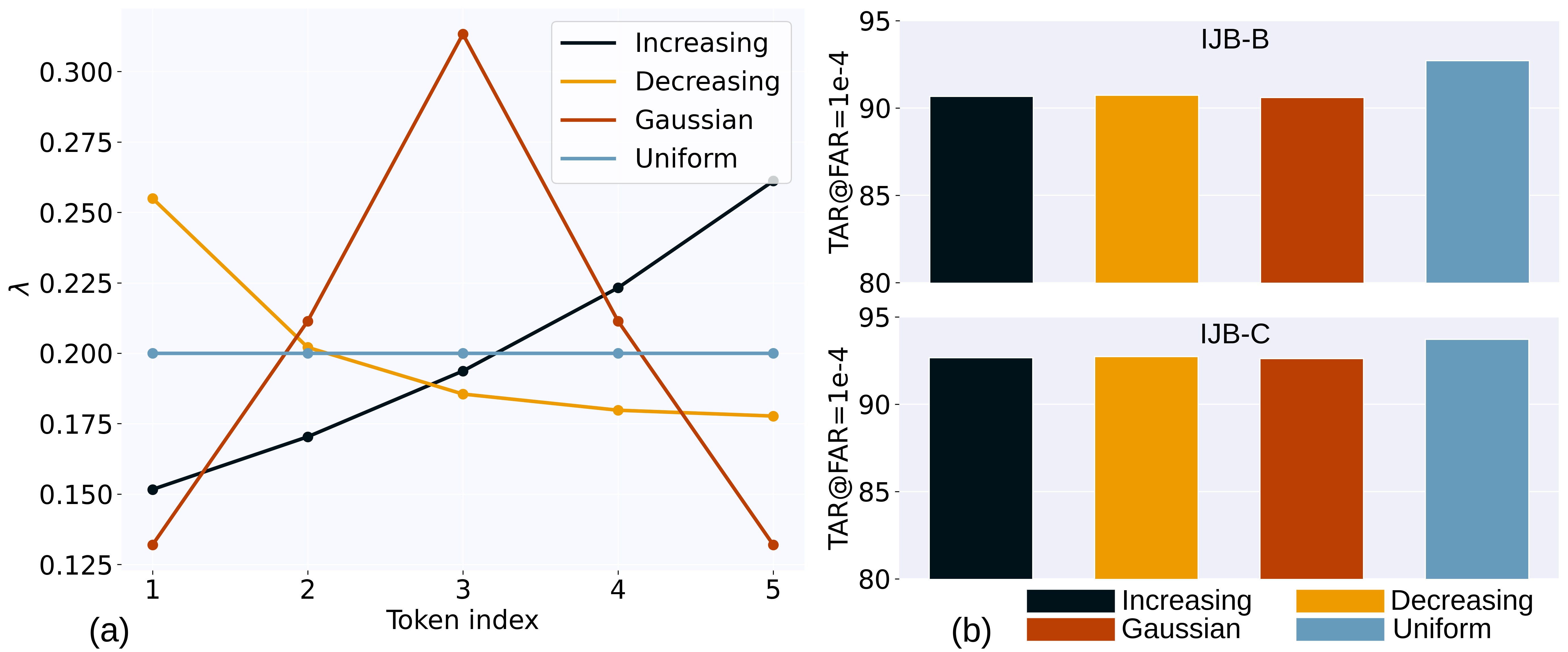}
\caption{a) Showing the value of the $\lambda$ for each token index in different scenarios. b) GIF performance is the best when the balancing factor of tokens, \ie, $\lambda$, is uniform across tokens.}\label{abblambda}
\end{figure}

\begin{figure}[t]
\centering
\includegraphics[width=1.0\linewidth]{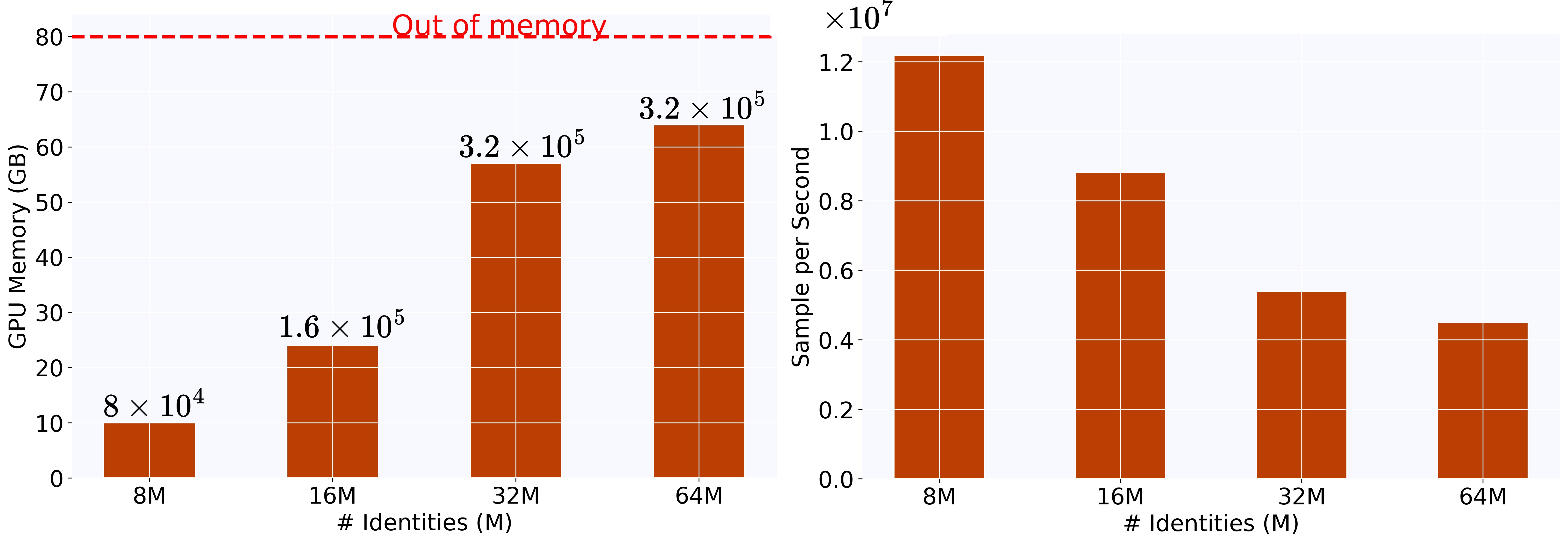}
\caption{GPU memory consumption (a) and Training speed (b) needed for optimization of code vectors.}\label{centercompute}
\vspace{-16pt}
\end{figure}

\section{Code Vector Optimization Cost}
\label{ccc}
Here, we investigate the GPU memory consumption associated with the optimization of code vectors. As demonstrated in Figure \ref{centercompute}a, even with the number of identities reaching 64 million, the GPU memory usage remains significantly lower than the OOM threshold. 
Moreover, Figure \ref{centercompute}b shows the remarkable speed of the optimization in this optimization.
This substantial reduction in memory consumption, coupled with improvements in processing speed and batch size, can be attributed to the fact that code vector optimization does not depend on the dataset or backbone architecture. Consequently, the optimization in Equation \ref{practical_loss} bypasses the time-intensive tasks of image loading and executing feedforward and backward passes through the backbone.

\section{Replacing CLIP with DINO}
\label{otclip}
In this study, we explore the sensitivity of GIF to changes in the model used for initializing code vectors. We conduct experiments employing the DINO \cite{caron2021emerging} representation, adjusting our embedding dimension to \(d=718\) to accommodate this model.
Results presented in Table \ref{clipVsDino} confirm our expectations: GIF demonstrates robustness to the specific pretrained model used to initialize the code vectors. Models trained on large datasets with the objective of developing a generalized representation, such as CLIP or DINO, prove adequate for initializing the code vectors since the proposed method solely needs the meaningful order of similarity from initialized code vectors.

\begin{table}
\addtolength{\tabcolsep}{-5pt}
\resizebox{1.0\linewidth}{!}{
\begin{tabular}{l|c|ccccccc}
\toprule
Method               & Train Set & LFW   & CPLFW & CALFW & CFP-FP & Age-DB & IJB-B & IJB-C \\ \midrule
GIF (DINO)            & MS1MV2            & \textbf{99.85} &        \textbf{94.47}               & 96.75                     & 98.75  &              \textbf{98.67}  & 94.98 & \textbf{96.80} \\
GIF (CLIP)  &   MS1MV2            &    \textbf{99.85}   & {94.45}      &  \textbf{96.94}     &  \textbf{98.80}      &     {98.58}   &   \textbf{95.05}    &   {96.77}    \\ \midrule
GIF (DINO)            & WebFace4M     & 99.83 & 94.97 & \textbf{96.92} & \textbf{98.41}  & \textbf{98.63}  & \textbf{96.95} & 97.76 \\
GIF  (CLIP)    & WebFace4M             &  \textbf{99.85}     &    \textbf{95.03}   & {96.85}      & {98.36}       &  \textbf{98.55}      &   {96.90}    & \textbf{97.83}      \\  \bottomrule
\end{tabular}}
\caption{Performance comparison to when we substitute  CLIP with DINO for initializing the code vectors. Verification accuracy (\%) is reported for LFW, CFP-FP, and AgeDB. TAR@FAR$=1e-4$ is reported for IJB-B and IJB-C.} \label{clipVsDino}
\vspace{-10pt}
\end{table}


\end{document}